\documentclass[sigconf]{aamas} 
\pdfoutput=1

\usepackage{balance} 
\usepackage{tikz} \usetikzlibrary{automata, positioning} 
\usepackage{pgfplots} 
\pgfplotsset{compat=1.18}
\usepackage[T1]{fontenc}
\makeatletter
\gdef\@copyrightpermission{
  \begin{minipage}{0.2\columnwidth}
   \href{https://creativecommons.org/licenses/by/4.0/}{\includegraphics[width=0.90\textwidth]{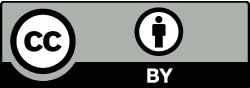}}
  \end{minipage}\hfill
  \begin{minipage}{0.8\columnwidth}
   \href{https://creativecommons.org/licenses/by/4.0/}{This work is licensed under a Creative Commons Attribution International 4.0 License.}
  \end{minipage}
  \vspace{5pt}
}
\makeatother

\setcopyright{ifaamas}
\acmConference[AAMAS '25]{Proc.\@ of the 24th International Conference
on Autonomous Agents and Multiagent Systems (AAMAS 2025)}{May 19 -- 23, 2025}
{Detroit, Michigan, USA}{Y.~Vorobeychik, S.~Das, A.~Nowé  (eds.)}
\copyrightyear{2025}
\acmYear{2025}
\acmDOI{}
\acmPrice{}
\acmISBN{}

\acmSubmissionID{bs080}

\title[AAMAS-2025 Formatting Instructions]{Grounding Agent Reasoning in Image Schemas: A Neurosymbolic Approach to Embodied Cognition}

\subtitle{Blue Sky Ideas Track}

\author{François Olivier}
\affiliation{
  \institution{CRIL CNRS \& Artois University}
  \city{Lens}
  \country{France}}
\email{olivier@cril.fr}

\author{Zied Bouraoui}
\affiliation{
  \institution{CRIL CNRS \& Artois University}
  \city{Lens}
  \country{France}}
\email{bouraoui@cril.fr}

\begin{abstract}
Despite advances in embodied AI, agent reasoning systems still struggle to capture the fundamental conceptual structures that humans naturally use to understand and interact with their environment. To address this, we propose a novel framework that bridges embodied cognition theory and agent systems by leveraging a formal characterization of image schemas, which are defined as recurring patterns of sensorimotor experience that structure human cognition. By customizing LLMs to translate natural language descriptions into formal representations based on these sensorimotor patterns, we will be able to create a neurosymbolic system that grounds the agent's understanding in fundamental conceptual structures. We argue that such an approach enhances both efficiency and interpretability while enabling more intuitive human-agent interactions through shared embodied understanding. 
\end{abstract}


\newcommand{\BibTeX}{\rm B\kern-.05em{\sc i\kern-.025em b}\kern-.08em\TeX}

\newcommand{\nextF}{\text{\rm \raisebox{-.5pt}{\Large\textopenbullet}}}
\newcommand{\alwaysF}{\ensuremath{\square}}
\newcommand{\eventuallyF}{\ensuremath{\lozenge}}
\newcommand{\final}{\ensuremath{\mathbb{F}}}

\newcommand{\untilF}{\ensuremath{\mathbb{U}}}

\newcommand{\eventuallyP}{\ensuremath{\blacklozenge}}

\begin{document}


\pagestyle{fancy}
\fancyhead{}


\maketitle 


\section{Introduction}
By the end of the 20th century, the classical paradigm of cognitive science was fundamentally challenged as evidence mounted that our minds do not operate like isolated symbol processing computers, but rather are inextricably linked to our bodily experiences in the world. This became particularly evident in how we understand and use language, as Lakoff and Johnson's groundbreaking work in `\textit{Metaphors We Live By}' \cite{lakoff1980metaphors} demonstrated that we comprehend abstract concepts (the target domain) by relying on our physical experiences as a source domain – we understand time through location ("the future is ahead of us"), importance through size ("this is a big deal"), and emotional states through spatial orientation ("I'm feeling down").

To bridge the gap between bodily experience and thought, Johnson \cite{johnson1987body} introduced \textit{image schemas} - recurring patterns abstracted from our sensorimotor interactions - and showed their pervasive role in structuring human thought across both concrete and abstract domains. Over the years, the theory has received robust experimental confirmation across multiple studies \cite{richardson2001language, mandler2014defining} and has proven fruitful even in non-linguistic domains such as mathematics \cite{lakoff2000mathematics}. A common example of image schema is \textsc{OBJECT\_INTO\_CONTAINER} which arises from our early physical experiences of putting objects into containers (e.g., cups and buckets), and later serves as a source domain to understand literal sentences like "Bill is in the house", more abstract ones such as "Berlin is in Germany" or "to be in love", and mathematical expressions such as "$ 2 \in \mathbb{N}$". More recent work has explored how these image schemas can be decomposed into even more basic constituents called \textit{conceptual primitives} \cite{mandler2014defining}. For instance, our comprehension of the concept of SUPPORT requires one to have the conceptual primitives of \textsc{UP/DOWN} and \textsc{CONTACT}. 

Just as cognitive science had to move beyond purely computational models to explain human cognition and linguistic ability, there is an ongoing debate about whether AI systems need similar grounding to achieve genuine language understanding and commonsense reasoning \cite{bender2020climbing, sogaard2023grounding}. While some recent work suggests that Large Language Models (LLMs) can grasp physical concepts through text alone \cite{patel2022mapping}, there are reasons to be skeptical about whether this statistical learning can capture the full depth of human conceptual understanding \cite{mccoy2023embers, MAHOWALD2024517}. For instance, \cite{DBLP:journals/corr/abs-2311-08993} highlights that LLMs employing in-context learning face significant challenges with tasks that require extensive specification, particularly those where even human annotators must carefully review a complex set of annotation guidelines to perform the task correctly. Using a simulation task, \cite{tamari-etal-2020-language} also demonstrates fundamental conceptual limitations of statistical methods - limitations that persist regardless of the scale of the data.
Equipping artificial agents with such conceptual embodied structures therefore becomes a crucial goal, as it would not only enable more intuitive and explainable human-agent interactions through shared embodied understanding, but also possibly represent, as suggested by \cite{bisk-etal-2020-experience}, the necessary step to move AI into its next major paradigm beyond current multimodal systems.

However, the primary challenge in achieving such agents is to formalize these psychological theories and deeply embodied structures, and to intertwine the resulting symbolic language with neural recognition and metaphorical mapping techniques in a promising way. In this work, we discuss the main challenges of such an endeavor and propose a promising approach that combines symbolic languages with neural architectures to create an integrated neurosymbolic framework. The main strengths of our approach compared to the existing work are the fully formal characterizations of conceptual structures, the use of existing symbolic solvers to reason with these characterizations, and the deep integration in a neural network in order to create a neurosymblic architecture. 

The rest of the paper is organized as follows. Section \ref{rel_work} presents some related work from a symbolic and machine learning perspective. Section \ref{prop} discusses some of the main properties that the expected formalism should satisfy in order to, as shown in Section \ref{form}, effectively capture the different conceptual primitives that compose image-schematic structures. Section \ref{parsing} presents how the formalism can be combined with neural networks in a meaningful way in order to enable fully embodied agents. Section \ref{nlu} discusses the advantages gained in reasoning and natural language understanding with such embodied agents. Section \ref{concl} concludes the paper.

\section{Related Work} \label{rel_work}
The formalization of image schemas is not a new endeavor - by the end of the twentieth century, Frank and Raubal \cite{frank1999formal} had already surveyed the existing formalisms. Among the formalisms that followed, notable approaches include bigraph-based representations \cite{amant2006image}, methods leveraging the WordNet lexical database \cite{kuhn2007image}, and approaches based on qualitative calculi \cite{bennett2014corpus, hedblom2020image}. Qualitative calculi, which generally correspond to relation algebras \cite{dylla2017survey}, appeared particularly well-suited for this formalization task, as they abstract away from precise numerical measurements similarly to how human cognitive processing focuses on relative relationships. Significantly advancing the field, Hedblom's work made extensive use of the adequacy of qualitative calculi by combining the Region Connection Calculus, Qualitative Trajectory Calculus, Cardinal Directions, and Linear Temporal Logic in order to represent both spatial and temporal dimensions of image schemas \cite{hedblom2020image}. Recently, Hedblom et al. proposed the Diagrammatic Image Schema Language (DISL) \cite{hedblom2024diagrammatic}, a systematic diagrammatic representation language for image schemas that provides a structured visual framework. 

Regarding the study of image schemas and embodied approaches in the machine learning community, the work of Wachowiak et al. explores how artificial agents capture implicit human intuitions underlying language \cite{wicke-wachowiak-2024-exploring} and introduces systematic methods for classifying natural language expressions into image schemas \cite{wachowiak-gromann-2022-systematic}. Recent advances in LLMs have also been leveraged to enhance performance in embodied learning tasks, particularly in Embodied Instruction Following \cite{shi-etal-2024-opex}, while standardized benchmarks to systematically evaluate these capabilities are emerging \cite{li2025embodied}. Finally, the framework developed in \cite{tamari-etal-2020-language} closely aligns with our goal by approaching language understanding through mental simulation and metaphoric mappings.

\section{Formalism Properties} \label{prop}
As initiated in \cite{mandler2014defining}, image schemas can be decomposed into conceptual primitives. For instance, \textsc{GOING\_IN} requires at least the notions of \textsc{OBJECT}, \textsc{CONTAINER} and \textsc{PATH}. To present our approach, we use the more recent classification from \cite{hedblom2024diagrammatic} reproduced in Table \ref{primitives_DISL}. As can be seen, some conceptual primitives are only spatial or spatiotemporal, whereas others are force dynamic primitives, which correspond to embodied feelings that cannot be represented in a spatiotemporal way (e.g., \textsc{UMPH} corresponds to the application of a \textit{force}).  

\begin{table}[h]
	\caption{Classification of conceptual primitives from \cite{hedblom2024diagrammatic}.}
	\label{primitives_DISL}
	\footnotesize  
	\begin{tabular}{llll}\toprule
		& \small entity & \small relational & \small attributive \\ \midrule
		\textbf{spatial} & OBJECT & LOCATION & OPEN \\
		& CONTAINER & START\_PATH  & CLOSED \\
		& PATH & END\_PATH   & EMPTY \\
		& REGION & CONTACT  & OCCUPIED \\
		& DOWN (/UP) & CONTAINED & FULL \\
		&  &SMALLER(/LARGER) & \\
		&  & PART\_OF & \\ \midrule
		\textbf{spatio-temporal} &  & PERMANENCE & MOTION \\
		& & & AT\_REST  \\
		& & & \tiny ANIMATE\_MOTION \\
		& & & \tiny INANIMATE\_MOTION \\ \midrule
		\textbf{force dynamic} &  & LINK& active-UMPH \\
		& &  &passive-UMPH \\ \bottomrule
	\end{tabular}
\end{table}

\noindent\textbf{Property 1.} Since image schemas can structure an infinite variety of physical configurations and scenarios, any formalism for representing them must be able to encode relationships qualitatively (e.g., being `inside' something without knowing exact locations or shapes) \cite{ligozat2013qualitative}. This requirement has been widely recognized in previous formalization attempts.

\noindent\textbf{Property 2.} Objects of different types can be involved in an image schema, such as points for atomic \textsc{OBJECTS} or lines for \textsc{PATHS}. Additionally, an ordering over types may be useful for defining certain entities (e.g., a \textsc{CONTAINER} can be a circle, a square, etc). Therefore, the formalism should be order-sorted and support the definition of typed relations. 

\noindent\textbf{Property 3.} Since image schemas can be understood as small narratives, the formalism should support the expression of time and the evolution of configurations over time. 

\noindent\textbf{Property 4.} The formalism should support quantification to express general rules and assert the (non-)existence of objects (e.g., for the primitive \textsc{EMPTY}), as well as logical connectives to effectively express logical constraints.

\noindent\textbf{Property 5.} Finally, the formalism should support the use of a default operator to model default behaviors, such as gravity or the law of inertia (i.e., things remain the same unless an action caused them to change) \cite{shanahan1997solving}. Importantly, the inclusion of a default operator makes the formalism non-monotonic.

\section{Formalizing Image Schemas} \label{form}
A promising candidate that meets these requirements, or allows for additional extensions to fulfill them, is to implement the Declarative Spatial Reasoning framework (DSR) \cite{DBLP:conf/cosit/BhattLS11} within the non-monotonic Quantified Equilibrium Logic with evaluable functions \cite{cabalar2011functional,cabalar2018functional}. Quantified equilibrium logic maintains the syntax of first-order logic while semantically interpreting negation as default negation (i.e., \textit{negation as failure} \cite{clark1977negation}).

\begin{figure}[h]
	\begin{tabular}{p{3.3cm}l}
            \begin{tikzpicture}
			 	\begin{axis}[
			 		width=4.4cm, height=4.4cm,
			 		axis x line=center, axis y line=center,
			 		axis line style={-}, xmin=0, xmax=10, ymin=0, ymax=10,
			 		xtick={0,...,10}, ytick={0,...,10},
			 		extra x ticks={0}, grid=both, grid style={gray!50},
			 		font=\small, extra y ticks={0}]
			 		\draw [thick](5,5) circle [radius=3] node [text=black, above=0.8cm]  {$c$};
			 		\fill[gray!10, opacity=0.7] (5,5) circle [radius=3];
			 		\draw [thick](6,4.5) circle [radius=1] node [text=black, above =0.25cm]  {$b$};
			 		\draw [fill=black](4,5) circle [radius=0.18] node [text=black, above=0.01cm]  {$a$};
			 	\end{axis}
			 \end{tikzpicture} 
		&
		\begin{tabular}[b]{lll}
		$x_a = 4$ & ~$x_b = 6$ & ~$x_c = 5$ \\ 	
		$y_a = 5$ & ~$y_b = 4.5$  &~ $y_c = 5$\\ 	
		                  & ~$r_b = 1$  & ~$r_c = 3$  \\
		                  && \\
		          \multicolumn{2}{l}{$inside(a, c) $  } & \\
		   		\multicolumn{3}{l}{$\leftrightarrow ~(x_a-x_c)^2+(y_a-y_c)^2 < r_c^2 $  }     \\ 
		   		\multicolumn{3}{l}{$\leftrightarrow ~(-1)^2 + 0^2 < 3^2 $  }     \\ 
		   		\multicolumn{3}{l}{$\leftrightarrow ~1 + 0 < 9 $  } \\
		   		\multicolumn{3}{l}{$\leftrightarrow ~ $ TRUE}\\
		\end{tabular}
		\\
	\end{tabular}
	\caption{In the DSR framework, the parameters of objects are used to define spatial relations between these objects.}
    \Description{A figure illustrating the DSR (Diagrammatic Spatial Reasoning) framework with two parts side by side. The left part shows a Cartesian coordinate system with a grid from 0 to 10 on both axes. It contains three elements: a point labeled 'a' at coordinates (4,5), a small circle labeled 'b' with radius 1 centered at (6,4.5), and a larger circle labeled 'c' with radius 3 centered at (5,5). Point 'a' is visibly inside circle 'c'. The right part shows the mathematical formulation of the spatial relation 'inside(a,c)' with the coordinates and radii of each element, followed by a step-by-step verification proving that point 'a' is indeed inside circle 'c'.}
    \label{DSR_examples}
\end{figure}
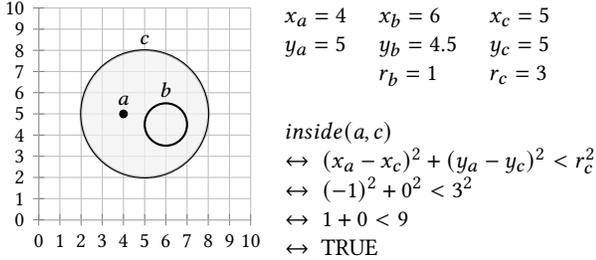

Evaluable functions enable the embedding of the DSR framework \cite{DBLP:conf/cosit/BhattLS11} since the latter fundamentally relies on parametric functions for representing objects (see Figure \ref{DSR_examples}, top right), and defines qualitative relations between objects through polynomial constraints on these parameters (bottom right) \cite{preparata2012computational}. Contrary to the common practice of using the algebraic qualitative calculi mentioned in Section~\ref{rel_work}, the DSR framework allows the combination of heterogeneous objects and does not impose any conditions on the set of relations defined.

Regarding temporal modeling, a first-order extension of Temporal Equilibrium Logic has been proposed in \cite{aguado2017temporal}. For the purpose of forthcoming examples, we consider the following temporal operators: 
\begin{center}
\begin{tabular}{rlcrlcrl}
		$\nextF$ & ~ \textit{next}&~~~~~~&$\alwaysF$ & ~ \textit{always} &~~~~~~&$\eventuallyF$ & ~ \textit{eventually afterward}\\
		$\untilF$ & ~ \textit{until} &~~~~~~& $\final$ & ~ \textit{final}&~~~~~~&$\eventuallyP$ & ~\textit{eventually before} \\
\end{tabular} 
\end{center}

\noindent Finally, many-sorted formalisms have been developed for approaches closely related to equilibrium logic \cite{balai2013towards}, while formal treatments of order-sorted logic can be found in \cite{kaneiwa2004order}.

In what follows, we explain how conceptual primitives are handled in our formalism and provide examples of some of their combinations. Our treatment shares similarities with \cite{hedblom2024diagrammatic} as we apply our formalism on their classification presented in Figure \ref{primitives_DISL}. 

Entities correspond to constants in the logic. The entity \textsc{OBJECT} simply corresponds to a point. The entity \textsc{CONTAINER} corresponds to any geometric objects that can be used in a relation of `containment', such as \textit{inside}, \textit{properPart}, etc. Ordered sorts enable us to define this entity as a superclass, that is, any circle, rectangle, etc. is a \textsc{CONTAINER} entity. The \textsc{PATH} entity is modeled as a line with a starting and ending point. For instance, the image schema of \textsc{SOURCE\_PATH\_GOAL} which underlies our understanding of processes composed of consecutive steps (e.g., the progression of degrees in a student's academic journey, advancing through the bases in baseball, etc), can be represented by a series of location such as $l_1 \land \eventuallyF(l_i \land \eventuallyF(... \land \eventuallyF l_n))$ where $l_1$ and $l_n$ respectively stand for the \textsc{START\_PATH} and END\_PATH as specific locations, and each $l_i$ represents an intermediate location. Forward movement is obtained by constraining the actual LOCATION with the previous ones by means of the $\eventuallyP$ operator. The entity REGION is modeled either by means of a distance function $\Delta$ or as a \textsc{CONTAINER} entity similar to the one above. Finally, the more abstract notion \textsc{DOWN} is either modeled as a line placed at the bottom of the scene, or is directly encoded within displacement actions. For instance, gravity can be modeled as $\alwaysF (\forall x (\neg \exists y ~on(x, y) \to moveDown(x)))$, where $x$ and $y$ are any entities in the domain. Note the use of the default negation in the latter formula. 

The relational primitives mainly correspond to binary (or higher-arity) relations. \textsc{LOCATION} can be expressed by means of positional or topological relations (e.g., \textit{on}, \textit{closeTo}, \textit{inside},...). \textsc{START\_PATH} and \textsc{END\_PATH}, as mentioned above, can be defined as points or geometric regions that delimitate a PATH entity. CONTACT, CONTAINED and PART\_OF simply correspond to topological relations defined in the DSR framework, and similarly for SMALLER/LARGER as size relations. LINK can either be defined by means of a distance $\Delta$ that cannot exceed a certain threshold or as an actual segment that touches the objects linked. Finally, PERMANENCE can be expressed by means of a default negation, encoding the idea that if we cannot prove that a parametric function of an entity has changed, we conserve its value for the actual state.

Although attributive conceptual primitives first seem to correspond to unary predicates applied to entities, they will usually require complex formulas. For instance, \textsc{EMPTY} corresponds to a formula where we state that, for a CONTAINER, no entity is inside it. The force dynamic conceptual primitives \textsf{active}\textsc{-UMPH} and \textsf{passive}\textsc{-UMPH} are modeled with default negation. Basically, \textit{unless} a contrary force is applied to an object, the latter is subject to an action at each state (possibly until a certain goal is achieved, using the $\untilF$ operator). Such a concept of force occurs in the characterization of gravity as presented above. Finally, \textsc{MOTION}, \textsc{AT\_REST} and the \textsc{(IN)ANIMATE} primitives correspond to action predicates that modify/apply to the location of entities along states. 

When combined, these conceptual primitives give rise to image schemas, each of which formally corresponds to a small theory $\Gamma$ encapsulating its essential structure and enabling reasoning. Such fully formal characterizations may also contribute to clarifying and standardizing the definitions of image schemas within the field. From a model-theoretic perspective, each model of a theory $\Gamma$ represents a possible instantiation of the structure under consideration, which aligns with the idea of a \textit{schema} used as a template for generating infinitely many concrete \textit{images} and scenarios. 
    
\section{Natural Language Parsing via Neural Image Schema Recognition} \label{parsing}
Having established a formal foundation for representing image schemas in the previous sections, we now turn to the challenge of automatically extracting these representations from natural language. Our goal is to develop a system that can take ordinary sentences and parse them into the non-monotonic quantified formalism presented above.

This task presents unique challenges compared to traditional semantic parsing. While conventional semantic parsers typically map language to classical logical systems \cite{pan2023logic, yang-etal-2024-harnessing}, our system must capture the embodied, spatiotemporal meaning inherent in language. For instance, when processing a sentence like "The monk climbs up the mountain" from the riddle presented in \cite{hedblom2024diagrammatic}, the system must recognize not only the entities involved but also the complex interplay of image schemas such as SOURCE\_PATH\_GOAL and CONTACT, along with their temporal evolution.

To address this challenge, we propose leveraging recent advances in LLMs and neural architectures. Modern transformer-based models have demonstrated remarkable capabilities in understanding linguistic structure and generating complex outputs. We can build on their strong language understanding and generation capabilities to translate natural language descriptions into our image schema formalism. A critical challenge in developing such a system is collecting sufficient high-quality mapping data between natural language sentences and their image schema representations. Fortunately, several existing resources can be leveraged: 
\begin{itemize}
\item Structured databases from \cite{wachowiak-gromann-2022-systematic, wicke-wachowiak-2024-exploring} provide ready-to-use examples for training, validation and testing.
\item Psychological experiments in the literature, for instance \cite{richardson2001language}, offer empirically-grounded data on image schema evocation in human participants.
\item LLMs can be strategically prompted to generate candidate image schema annotations for natural language sentences.
\item Expert linguists and cognitive scientists can provide gold-standard annotations mapping linguistic constituents to schema roles and identifying active image schemas.
\end{itemize}

\balance 
Regarding the formalization of the image schema representations, we propose a two-stage approach. First, we can leverage LLMs' strong reasoning capabilities to generate initial formal characterizations of identified schemas. Our formalism's adherence to first-order logic with temporal operators makes it particularly amenable to automated generation, as these logical structures are well-represented in LLMs' training data. Second, we can fine-tune a specialized translation model on our collected dataset of natural language sentences paired with their formal representations. This model would learn to directly map input text to well-formed expressions in our formalism.
To ensure quality and consistency, we propose an iterative development process where model outputs are validated against expert annotations and refined based on error analysis. 

Finally, evaluation of such a system requires going beyond simple accuracy metrics. While an exact match with gold-standard annotations provides one measure of success, we must also consider partial matching metrics that assess the system's ability to identify correct image schemas, assign appropriate roles, and maintain proper temporal structures. Additionally, the system's performance should be evaluated on downstream tasks that require genuine understanding of spatial relationships, motion events and force components.

\section{Natural Language Understanding, Reasoning and Analogies} \label{nlu}
The proposed model could serve as a crucial component in embodied AI systems, helping to bridge the gap between language understanding and physical interaction with the world. Image schemas, being grounded in bodily experience and spatial understanding, provide a natural intermediate representation between linguistic input and physical action. By capturing these embodied cognitive patterns in our formal notation, we enable AI systems to process language in a way that connects directly to spatial reasoning and motor planning. This creates a tighter coupling between natural language understanding and real-world interaction - rather than treating language as purely symbolic manipulations, the system can ground linguistic meanings in the same kind of spatial and motor primitives that humans use.

Reasoning would also be enhanced through closer alignment with human cognitive processes. By operating over the same kind of image-schematic representations that humans use, AI systems could better model and predict human understanding and misunderstanding. For example, an agent could identify when a human might struggle to grasp a concept by analyzing which image schemas are involved and whether they map naturally to familiar embodied experiences. Moreover, these agents could reason in ways that parallel human inference patterns. As Shimojima demonstrates in his analysis of diagrammatic reasoning \cite{Shimojima2015-SHISPO-2}, certain conclusions emerge naturally (or come "for free") from visual representations without explicit logical rules. Image schemas leverage this same principle, as the spatial constraints between entities capture the logical constraints in the target domain \cite{olivier:tel-03984759}. To realize these inferences in a computational framework, we can harness answer set programming via Clingo, as partly explored in work on related areas \cite{walega_schultz_bhatt_2017, schultz2018answer, DBLP:journals/corr/SuchanBJ15}. Clingo's ability to handle non-monotonic reasoning and incorporate custom theories such as the ones described to characterize image schemas makes it particularly suitable for implementing our formalism. 

Finally, our formalism might turn out particularly useful in capturing analogical relationships, where a conceptual structure can be mapped to multiple target domains. Consider the classic analogy between the solar and (Rutherford–Bohr) atomic systems, exemplified in the sentences "electrons circle the nucleus" and "planets circle the sun" \cite{jiayang2023storyanalogy}. Both can be formalized using the same image-schematic structure where a distance $\Delta(x,y)$, between $x$ as electrons/planets and $y$ as the nucleus/sun, is constrained within certain bounds, and $\theta(x,y) < \nextF \theta(x,y)$ ensures that the angular position of $x$ relative to $y$ continuously increases, capturing the circular orbital motion. The structural similarity revealed in these formalizations explains the cognitive power of the analogy - both scenarios share the same underlying image-schematic structure.

\section{Conclusion and Challenges} \label{concl}
This paper has presented a comprehensive approach to bridging the gap between natural language understanding and embodied cognition. Building on cognitive theories of image schemas and recent advances in large language models, we have outlined a formalism that captures the essential spatial, temporal and force dynamic primitives underlying human conceptual understanding. While the complete formalization remains to be fully developed, we have demonstrated how the key components can be systematically combined to represent complex conceptual structures. The integration of this formalism with modern transformer architectures opens new possibilities for grounding language understanding in embodied experiences. By capturing image schemas in a computationally tractable form, we enable systems to process language in ways that mirror human cognitive patterns. The resulting representations support natural forms of reasoning and analogical mapping, as demonstrated through examples ranging from basic containment relationships to complex analogies. Our work provides a foundation for developing AI systems that can understand and reason with language in more human-like ways. 



\begin{acks}
This work was supported by the French National Research Agency (ANR) under grant ANR-22-CE23-0002 ERIANA.
\end{acks}



\bibliographystyle{ACM-Reference-Format} 
\bibliography{aamas_2025_Olivier_Bouraoui}


\begin{thebibliography}{42}


\ifx \showCODEN    \undefined \def \showCODEN     #1{\unskip}     \fi
\ifx \showDOI      \undefined \def \showDOI       #1{#1}\fi
\ifx \showISBNx    \undefined \def \showISBNx     #1{\unskip}     \fi
\ifx \showISBNxiii \undefined \def \showISBNxiii  #1{\unskip}     \fi
\ifx \showISSN     \undefined \def \showISSN      #1{\unskip}     \fi
\ifx \showLCCN     \undefined \def \showLCCN      #1{\unskip}     \fi
\ifx \shownote     \undefined \def \shownote      #1{#1}          \fi
\ifx \showarticletitle \undefined \def \showarticletitle #1{#1}   \fi
\ifx \showURL      \undefined \def \showURL       {\relax}        \fi
\providecommand\bibfield[2]{#2}
\providecommand\bibinfo[2]{#2}
\providecommand\natexlab[1]{#1}
\providecommand\showeprint[2][]{arXiv:#2}

\bibitem[\protect\citeauthoryear{Aguado, Cabalar, P{\'e}rez, Vidal, and
  Dieguez}{Aguado et~al\mbox{.}}{2017}]%
        {aguado2017temporal}
\bibfield{author}{\bibinfo{person}{Felicidad Aguado}, \bibinfo{person}{Pedro
  Cabalar}, \bibinfo{person}{Gilberto P{\'e}rez},
  \bibinfo{person}{Concepci{\'o}n Vidal}, {and} \bibinfo{person}{Martin
  Dieguez}.} \bibinfo{year}{2017}\natexlab{}.
\newblock \showarticletitle{Temporal logic programs with variables}.
\newblock \bibinfo{journal}{\emph{Theory and Practice of Logic Programming}}
  \bibinfo{volume}{17}, \bibinfo{number}{2} (\bibinfo{year}{2017}),
  \bibinfo{pages}{226--243}.
\newblock


\bibitem[\protect\citeauthoryear{Amant, Morrison, Chang, Cohen, and Beal}{Amant
  et~al\mbox{.}}{2006}]%
        {amant2006image}
\bibfield{author}{\bibinfo{person}{Robert~St Amant}, \bibinfo{person}{Clayton~T
  Morrison}, \bibinfo{person}{Yu-Han Chang}, \bibinfo{person}{Paul~R Cohen},
  {and} \bibinfo{person}{Carole Beal}.} \bibinfo{year}{2006}\natexlab{}.
\newblock \showarticletitle{An image schema language}. In
  \bibinfo{booktitle}{\emph{Submitted to The 7th International Conference on
  Cognitive Modelling (ICCM 2006)}}. Citeseer.
\newblock


\bibitem[\protect\citeauthoryear{Balai, Gelfond, and Zhang}{Balai
  et~al\mbox{.}}{2013}]%
        {balai2013towards}
\bibfield{author}{\bibinfo{person}{Evgenii Balai}, \bibinfo{person}{Michael
  Gelfond}, {and} \bibinfo{person}{Yuanlin Zhang}.}
  \bibinfo{year}{2013}\natexlab{}.
\newblock \showarticletitle{Towards Answer Set Programming with Sorts}. In
  \bibinfo{booktitle}{\emph{Proceedings of the 12th International Conference on
  Logic Programming and Nonmonotonic Reasoning - Volume 8148}} (Corunna, Spain)
  \emph{(\bibinfo{series}{LPNMR 2013})}. \bibinfo{publisher}{Springer-Verlag},
  \bibinfo{address}{Berlin, Heidelberg}, \bibinfo{pages}{135–147}.
\newblock
\showISBNx{9783642405631}
\urldef\tempurl%
\url{https://doi.org/10.1007/978-3-642-40564-8_14}
\showDOI{\tempurl}


\bibitem[\protect\citeauthoryear{Bender and Koller}{Bender and Koller}{2020}]%
        {bender2020climbing}
\bibfield{author}{\bibinfo{person}{Emily~M Bender} {and}
  \bibinfo{person}{Alexander Koller}.} \bibinfo{year}{2020}\natexlab{}.
\newblock \showarticletitle{Climbing towards NLU: On meaning, form, and
  understanding in the age of data}. In \bibinfo{booktitle}{\emph{Proceedings
  of the 58th annual meeting of the association for computational
  linguistics}}. \bibinfo{pages}{5185--5198}.
\newblock


\bibitem[\protect\citeauthoryear{Bennett and Cialone}{Bennett and
  Cialone}{2014}]%
        {bennett2014corpus}
\bibfield{author}{\bibinfo{person}{Brandon Bennett} {and}
  \bibinfo{person}{Claudia Cialone}.} \bibinfo{year}{2014}\natexlab{}.
\newblock \showarticletitle{Corpus Guided Sense Cluster Analysis: a methodology
  for ontology development (with examples from the spatial domain)}. In
  \bibinfo{booktitle}{\emph{Formal Ontology in Information Systems}}. IOS
  Press, \bibinfo{pages}{213--226}.
\newblock


\bibitem[\protect\citeauthoryear{Bhatt, Lee, and Schultz}{Bhatt
  et~al\mbox{.}}{2011}]%
        {DBLP:conf/cosit/BhattLS11}
\bibfield{author}{\bibinfo{person}{Mehul Bhatt}, \bibinfo{person}{Jae~Hee Lee},
  {and} \bibinfo{person}{Carl Schultz}.} \bibinfo{year}{2011}\natexlab{}.
\newblock \showarticletitle{{CLP(QS):} {A} Declarative Spatial Reasoning
  Framework}. In \bibinfo{booktitle}{\emph{Spatial Information Theory - 10th
  International Conference, {COSIT} 2011, Belfast, ME, USA, September 12-16,
  2011. Proceedings}} \emph{(\bibinfo{series}{Lecture Notes in Computer
  Science}, Vol.~\bibinfo{volume}{6899})},
  \bibfield{editor}{\bibinfo{person}{Max~J. Egenhofer},
  \bibinfo{person}{Nicholas~A. Giudice}, \bibinfo{person}{Reinhard Moratz},
  {and} \bibinfo{person}{Michael~F. Worboys}} (Eds.).
  \bibinfo{publisher}{Springer}, \bibinfo{pages}{210--230}.
\newblock
\urldef\tempurl%
\url{https://doi.org/10.1007/978-3-642-23196-4\_12}
\showDOI{\tempurl}


\bibitem[\protect\citeauthoryear{Bisk, Holtzman, Thomason, Andreas, Bengio,
  Chai, Lapata, Lazaridou, May, Nisnevich, Pinto, and Turian}{Bisk
  et~al\mbox{.}}{2020}]%
        {bisk-etal-2020-experience}
\bibfield{author}{\bibinfo{person}{Yonatan Bisk}, \bibinfo{person}{Ari
  Holtzman}, \bibinfo{person}{Jesse Thomason}, \bibinfo{person}{Jacob Andreas},
  \bibinfo{person}{Yoshua Bengio}, \bibinfo{person}{Joyce Chai},
  \bibinfo{person}{Mirella Lapata}, \bibinfo{person}{Angeliki Lazaridou},
  \bibinfo{person}{Jonathan May}, \bibinfo{person}{Aleksandr Nisnevich},
  \bibinfo{person}{Nicolas Pinto}, {and} \bibinfo{person}{Joseph Turian}.}
  \bibinfo{year}{2020}\natexlab{}.
\newblock \showarticletitle{Experience Grounds Language}. In
  \bibinfo{booktitle}{\emph{Proceedings of the 2020 Conference on Empirical
  Methods in Natural Language Processing (EMNLP)}},
  \bibfield{editor}{\bibinfo{person}{Bonnie Webber}, \bibinfo{person}{Trevor
  Cohn}, \bibinfo{person}{Yulan He}, {and} \bibinfo{person}{Yang Liu}} (Eds.).
  \bibinfo{publisher}{Association for Computational Linguistics},
  \bibinfo{address}{Online}, \bibinfo{pages}{8718--8735}.
\newblock
\urldef\tempurl%
\url{https://doi.org/10.18653/v1/2020.emnlp-main.703}
\showDOI{\tempurl}


\bibitem[\protect\citeauthoryear{Cabalar}{Cabalar}{2011}]%
        {cabalar2011functional}
\bibfield{author}{\bibinfo{person}{Pedro Cabalar}.}
  \bibinfo{year}{2011}\natexlab{}.
\newblock \showarticletitle{Functional answer set programming}.
\newblock \bibinfo{journal}{\emph{Theory and Practice of Logic Programming}}
  \bibinfo{volume}{11}, \bibinfo{number}{2-3} (\bibinfo{year}{2011}),
  \bibinfo{pages}{203--233}.
\newblock


\bibitem[\protect\citeauthoryear{Cabalar, Fandinno, Del~Cerro, and
  Pearce}{Cabalar et~al\mbox{.}}{2018}]%
        {cabalar2018functional}
\bibfield{author}{\bibinfo{person}{Pedro Cabalar}, \bibinfo{person}{Jorge
  Fandinno}, \bibinfo{person}{Luis~Farinas Del~Cerro}, {and}
  \bibinfo{person}{David Pearce}.} \bibinfo{year}{2018}\natexlab{}.
\newblock \showarticletitle{Functional ASP with intensional sets: Application
  to Gelfond-Zhang aggregates}.
\newblock \bibinfo{journal}{\emph{Theory and Practice of Logic Programming}}
  \bibinfo{volume}{18}, \bibinfo{number}{3-4} (\bibinfo{year}{2018}),
  \bibinfo{pages}{390--405}.
\newblock


\bibitem[\protect\citeauthoryear{Clark}{Clark}{1977}]%
        {clark1977negation}
\bibfield{author}{\bibinfo{person}{Keith~L Clark}.}
  \bibinfo{year}{1977}\natexlab{}.
\newblock \showarticletitle{Negation as failure}.
\newblock In \bibinfo{booktitle}{\emph{Logic and data bases}}.
  \bibinfo{publisher}{Springer}, \bibinfo{pages}{293--322}.
\newblock


\bibitem[\protect\citeauthoryear{Dylla, Lee, Mossakowski, Schneider, Delden,
  Ven, and Wolter}{Dylla et~al\mbox{.}}{2017}]%
        {dylla2017survey}
\bibfield{author}{\bibinfo{person}{Frank Dylla}, \bibinfo{person}{Jae~Hee Lee},
  \bibinfo{person}{Till Mossakowski}, \bibinfo{person}{Thomas Schneider},
  \bibinfo{person}{Andr{\'e}~Van Delden}, \bibinfo{person}{Jasper Van~De Ven},
  {and} \bibinfo{person}{Diedrich Wolter}.} \bibinfo{year}{2017}\natexlab{}.
\newblock \showarticletitle{A survey of qualitative spatial and temporal
  calculi: Algebraic and computational properties}.
\newblock \bibinfo{journal}{\emph{ACM Computing Surveys (CSUR)}}
  \bibinfo{volume}{50}, \bibinfo{number}{1} (\bibinfo{year}{2017}),
  \bibinfo{pages}{1--39}.
\newblock


\bibitem[\protect\citeauthoryear{Frank and Raubal}{Frank and Raubal}{1999}]%
        {frank1999formal}
\bibfield{author}{\bibinfo{person}{Andrew~U Frank} {and}
  \bibinfo{person}{Martin Raubal}.} \bibinfo{year}{1999}\natexlab{}.
\newblock \showarticletitle{Formal specification of image schemata--a step
  towards interoperability in geographic information systems}.
\newblock \bibinfo{journal}{\emph{Spatial Cognition and Computation}}
  \bibinfo{volume}{1} (\bibinfo{year}{1999}), \bibinfo{pages}{67--101}.
\newblock


\bibitem[\protect\citeauthoryear{Hedblom}{Hedblom}{2020}]%
        {hedblom2020image}
\bibfield{author}{\bibinfo{person}{Maria~M Hedblom}.}
  \bibinfo{year}{2020}\natexlab{}.
\newblock \bibinfo{booktitle}{\emph{Image schemas and concept invention:
  cognitive, logical, and linguistic investigations}}.
\newblock \bibinfo{publisher}{Springer Nature}.
\newblock


\bibitem[\protect\citeauthoryear{Hedblom, Neuhaus, and Mossakowski}{Hedblom
  et~al\mbox{.}}{2024}]%
        {hedblom2024diagrammatic}
\bibfield{author}{\bibinfo{person}{Maria~M Hedblom}, \bibinfo{person}{Fabian
  Neuhaus}, {and} \bibinfo{person}{Till Mossakowski}.}
  \bibinfo{year}{2024}\natexlab{}.
\newblock \showarticletitle{The Diagrammatic Image Schema Language (DISL)}.
\newblock \bibinfo{journal}{\emph{Spatial Cognition \& Computation}}
  (\bibinfo{year}{2024}), \bibinfo{pages}{1--38}.
\newblock


\bibitem[\protect\citeauthoryear{Jiayang, Qiu, Chan, Fang, Wang, Chan, Ru, Guo,
  Zhang, Song, et~al\mbox{.}}{Jiayang et~al\mbox{.}}{2023}]%
        {jiayang2023storyanalogy}
\bibfield{author}{\bibinfo{person}{Cheng Jiayang}, \bibinfo{person}{Lin Qiu},
  \bibinfo{person}{Tsz~Ho Chan}, \bibinfo{person}{Tianqing Fang},
  \bibinfo{person}{Weiqi Wang}, \bibinfo{person}{Chunkit Chan},
  \bibinfo{person}{Dongyu Ru}, \bibinfo{person}{Qipeng Guo},
  \bibinfo{person}{Hongming Zhang}, \bibinfo{person}{Yangqiu Song},
  {et~al\mbox{.}}} \bibinfo{year}{2023}\natexlab{}.
\newblock \showarticletitle{StoryAnalogy: Deriving Story-level Analogies from
  Large Language Models to Unlock Analogical Understanding}.
\newblock \bibinfo{journal}{\emph{arXiv preprint arXiv:2310.12874}}
  (\bibinfo{year}{2023}).
\newblock


\bibitem[\protect\citeauthoryear{Johnson}{Johnson}{1987}]%
        {johnson1987body}
\bibfield{author}{\bibinfo{person}{Mark Johnson}.}
  \bibinfo{year}{1987}\natexlab{}.
\newblock \bibinfo{booktitle}{\emph{The body in the mind: The bodily basis of
  reason and imagination}}.
\newblock \bibinfo{publisher}{Chicago: University of Chicago Press}.
\newblock


\bibitem[\protect\citeauthoryear{Kaneiwa}{Kaneiwa}{2004}]%
        {kaneiwa2004order}
\bibfield{author}{\bibinfo{person}{Ken Kaneiwa}.}
  \bibinfo{year}{2004}\natexlab{}.
\newblock \showarticletitle{Order-sorted logic programming with predicate
  hierarchy}.
\newblock \bibinfo{journal}{\emph{Artificial Intelligence}}
  \bibinfo{volume}{158}, \bibinfo{number}{2} (\bibinfo{year}{2004}),
  \bibinfo{pages}{155--188}.
\newblock


\bibitem[\protect\citeauthoryear{Kuhn}{Kuhn}{2007}]%
        {kuhn2007image}
\bibfield{author}{\bibinfo{person}{Werner Kuhn}.}
  \bibinfo{year}{2007}\natexlab{}.
\newblock \showarticletitle{An image-schematic account of spatial categories}.
  In \bibinfo{booktitle}{\emph{International Conference on Spatial Information
  Theory}}. Springer, \bibinfo{pages}{152--168}.
\newblock


\bibitem[\protect\citeauthoryear{Lakoff and Johnson}{Lakoff and
  Johnson}{1980}]%
        {lakoff1980metaphors}
\bibfield{author}{\bibinfo{person}{George Lakoff} {and} \bibinfo{person}{Mark
  Johnson}.} \bibinfo{year}{1980}\natexlab{}.
\newblock \bibinfo{booktitle}{\emph{Metaphors We Live By}}.
\newblock \bibinfo{publisher}{University of Chicago Press}.
\newblock


\bibitem[\protect\citeauthoryear{Lakoff and N{\'u}{\~n}ez}{Lakoff and
  N{\'u}{\~n}ez}{2000}]%
        {lakoff2000mathematics}
\bibfield{author}{\bibinfo{person}{George Lakoff} {and} \bibinfo{person}{Rafael
  N{\'u}{\~n}ez}.} \bibinfo{year}{2000}\natexlab{}.
\newblock \bibinfo{booktitle}{\emph{Where mathematics comes from}}.
  Vol.~\bibinfo{volume}{6}.
\newblock \bibinfo{publisher}{New York: Basic Books}.
\newblock


\bibitem[\protect\citeauthoryear{Li, Zhao, Wang, Wang, Zhou, Srivastava,
  Gokmen, Lee, Li, Zhang, et~al\mbox{.}}{Li et~al\mbox{.}}{2025}]%
        {li2025embodied}
\bibfield{author}{\bibinfo{person}{Manling Li}, \bibinfo{person}{Shiyu Zhao},
  \bibinfo{person}{Qineng Wang}, \bibinfo{person}{Kangrui Wang},
  \bibinfo{person}{Yu Zhou}, \bibinfo{person}{Sanjana Srivastava},
  \bibinfo{person}{Cem Gokmen}, \bibinfo{person}{Tony Lee},
  \bibinfo{person}{Erran~Li Li}, \bibinfo{person}{Ruohan Zhang},
  {et~al\mbox{.}}} \bibinfo{year}{2025}\natexlab{}.
\newblock \showarticletitle{Embodied agent interface: Benchmarking llms for
  embodied decision making}.
\newblock \bibinfo{journal}{\emph{Advances in Neural Information Processing
  Systems}}  \bibinfo{volume}{37} (\bibinfo{year}{2025}),
  \bibinfo{pages}{100428--100534}.
\newblock


\bibitem[\protect\citeauthoryear{Ligozat}{Ligozat}{2013}]%
        {ligozat2013qualitative}
\bibfield{author}{\bibinfo{person}{G{\'e}rard Ligozat}.}
  \bibinfo{year}{2013}\natexlab{}.
\newblock \bibinfo{booktitle}{\emph{Qualitative spatial and temporal
  reasoning}}.
\newblock \bibinfo{publisher}{John Wiley \& Sons}.
\newblock


\bibitem[\protect\citeauthoryear{Mahowald, Ivanova, Blank, Kanwisher,
  Tenenbaum, and Fedorenko}{Mahowald et~al\mbox{.}}{2024}]%
        {MAHOWALD2024517}
\bibfield{author}{\bibinfo{person}{Kyle Mahowald}, \bibinfo{person}{Anna~A.
  Ivanova}, \bibinfo{person}{Idan~A. Blank}, \bibinfo{person}{Nancy Kanwisher},
  \bibinfo{person}{Joshua~B. Tenenbaum}, {and} \bibinfo{person}{Evelina
  Fedorenko}.} \bibinfo{year}{2024}\natexlab{}.
\newblock \showarticletitle{Dissociating language and thought in large language
  models}.
\newblock \bibinfo{journal}{\emph{Trends in Cognitive Sciences}}
  \bibinfo{volume}{28}, \bibinfo{number}{6} (\bibinfo{year}{2024}),
  \bibinfo{pages}{517--540}.
\newblock
\showISSN{1364-6613}
\urldef\tempurl%
\url{https://doi.org/10.1016/j.tics.2024.01.011}
\showDOI{\tempurl}


\bibitem[\protect\citeauthoryear{Mandler and C{\'a}novas}{Mandler and
  C{\'a}novas}{2014}]%
        {mandler2014defining}
\bibfield{author}{\bibinfo{person}{Jean~M Mandler} {and}
  \bibinfo{person}{Crist{\'o}bal~Pag{\'a}n C{\'a}novas}.}
  \bibinfo{year}{2014}\natexlab{}.
\newblock \showarticletitle{On defining image schemas}.
\newblock \bibinfo{journal}{\emph{Language and Cognition}} \bibinfo{volume}{6},
  \bibinfo{number}{4} (\bibinfo{year}{2014}), \bibinfo{pages}{510--532}.
\newblock


\bibitem[\protect\citeauthoryear{McCoy, Yao, Friedman, Hardy, and
  Griffiths}{McCoy et~al\mbox{.}}{2023}]%
        {mccoy2023embers}
\bibfield{author}{\bibinfo{person}{R~Thomas McCoy}, \bibinfo{person}{Shunyu
  Yao}, \bibinfo{person}{Dan Friedman}, \bibinfo{person}{Matthew Hardy}, {and}
  \bibinfo{person}{Thomas~L Griffiths}.} \bibinfo{year}{2023}\natexlab{}.
\newblock \showarticletitle{Embers of autoregression: Understanding large
  language models through the problem they are trained to solve}.
\newblock \bibinfo{journal}{\emph{arXiv preprint arXiv:2309.13638}}
  (\bibinfo{year}{2023}).
\newblock


\bibitem[\protect\citeauthoryear{Olivier}{Olivier}{2022}]%
        {olivier:tel-03984759}
\bibfield{author}{\bibinfo{person}{Fran{\c c}ois Olivier}.}
  \bibinfo{year}{2022}\natexlab{}.
\newblock \emph{\bibinfo{title}{{Spatial relations in reasoning : a
  computational model}}}.
\newblock Theses. \bibinfo{school}{{Universit{\'e} Paris sciences et lettres}}.
\newblock
\urldef\tempurl%
\url{https://theses.hal.science/tel-03984759}
\showURL{%
\tempurl}


\bibitem[\protect\citeauthoryear{Pan, Albalak, Wang, and Wang}{Pan
  et~al\mbox{.}}{2023}]%
        {pan2023logic}
\bibfield{author}{\bibinfo{person}{Liangming Pan}, \bibinfo{person}{Alon
  Albalak}, \bibinfo{person}{Xinyi Wang}, {and} \bibinfo{person}{William~Yang
  Wang}.} \bibinfo{year}{2023}\natexlab{}.
\newblock \showarticletitle{Logic-lm: Empowering large language models with
  symbolic solvers for faithful logical reasoning}.
\newblock \bibinfo{journal}{\emph{arXiv preprint arXiv:2305.12295}}
  (\bibinfo{year}{2023}).
\newblock


\bibitem[\protect\citeauthoryear{Patel and Pavlick}{Patel and Pavlick}{2022}]%
        {patel2022mapping}
\bibfield{author}{\bibinfo{person}{Roma Patel} {and} \bibinfo{person}{Ellie
  Pavlick}.} \bibinfo{year}{2022}\natexlab{}.
\newblock \showarticletitle{Mapping language models to grounded conceptual
  spaces}. In \bibinfo{booktitle}{\emph{International conference on learning
  representations}}.
\newblock


\bibitem[\protect\citeauthoryear{Peng, Wang, Chen, Li, Qi, Wang, Wu, Zeng, Xu,
  Hou, and Li}{Peng et~al\mbox{.}}{2023}]%
        {DBLP:journals/corr/abs-2311-08993}
\bibfield{author}{\bibinfo{person}{Hao Peng}, \bibinfo{person}{Xiaozhi Wang},
  \bibinfo{person}{Jianhui Chen}, \bibinfo{person}{Weikai Li},
  \bibinfo{person}{Yunjia Qi}, \bibinfo{person}{Zimu Wang},
  \bibinfo{person}{Zhili Wu}, \bibinfo{person}{Kaisheng Zeng},
  \bibinfo{person}{Bin Xu}, \bibinfo{person}{Lei Hou}, {and}
  \bibinfo{person}{Juanzi Li}.} \bibinfo{year}{2023}\natexlab{}.
\newblock \showarticletitle{When does In-context Learning Fall Short and Why?
  {A} Study on Specification-Heavy Tasks}.
\newblock \bibinfo{journal}{\emph{CoRR}}  \bibinfo{volume}{abs/2311.08993}
  (\bibinfo{year}{2023}).
\newblock
\urldef\tempurl%
\url{https://doi.org/10.48550/ARXIV.2311.08993}
\showDOI{\tempurl}
\showeprint[arXiv]{2311.08993}


\bibitem[\protect\citeauthoryear{Preparata and Shamos}{Preparata and
  Shamos}{2012}]%
        {preparata2012computational}
\bibfield{author}{\bibinfo{person}{Franco~P Preparata} {and}
  \bibinfo{person}{Michael~I Shamos}.} \bibinfo{year}{2012}\natexlab{}.
\newblock \bibinfo{booktitle}{\emph{Computational geometry: an introduction}}.
\newblock \bibinfo{publisher}{Springer Science \& Business Media}.
\newblock


\bibitem[\protect\citeauthoryear{Richardson, Spivey, Edelman, and
  Naples}{Richardson et~al\mbox{.}}{2001}]%
        {richardson2001language}
\bibfield{author}{\bibinfo{person}{Daniel~C Richardson},
  \bibinfo{person}{Michael~J Spivey}, \bibinfo{person}{Shimon Edelman}, {and}
  \bibinfo{person}{Adam~J Naples}.} \bibinfo{year}{2001}\natexlab{}.
\newblock \showarticletitle{" Language is spatial": Experimental evidence for
  image schemas of concrete and abstract verbs}. In
  \bibinfo{booktitle}{\emph{Proceedings of the Annual Meeting of the Cognitive
  Science Society}}, Vol.~\bibinfo{volume}{23}.
\newblock


\bibitem[\protect\citeauthoryear{Schultz, Bhatt, Suchan, and
  Wa{\l}{\k{e}}ga}{Schultz et~al\mbox{.}}{2018}]%
        {schultz2018answer}
\bibfield{author}{\bibinfo{person}{Carl Schultz}, \bibinfo{person}{Mehul
  Bhatt}, \bibinfo{person}{Jakob Suchan}, {and}
  \bibinfo{person}{Przemys{\l}aw~Andrzej Wa{\l}{\k{e}}ga}.}
  \bibinfo{year}{2018}\natexlab{}.
\newblock \showarticletitle{Answer Set Programming Modulo ‘Space-Time’}. In
  \bibinfo{booktitle}{\emph{International Joint Conference on Rules and
  Reasoning}}. Springer, \bibinfo{pages}{318--326}.
\newblock


\bibitem[\protect\citeauthoryear{Shanahan}{Shanahan}{1997}]%
        {shanahan1997solving}
\bibfield{author}{\bibinfo{person}{Murray Shanahan}.}
  \bibinfo{year}{1997}\natexlab{}.
\newblock \bibinfo{booktitle}{\emph{Solving the frame problem: a mathematical
  investigation of the common sense law of inertia}}.
\newblock \bibinfo{publisher}{MIT press}.
\newblock


\bibitem[\protect\citeauthoryear{Shi, Sun, Yuan, C{\^o}t{\'e}, and Liu}{Shi
  et~al\mbox{.}}{2024}]%
        {shi-etal-2024-opex}
\bibfield{author}{\bibinfo{person}{Haochen Shi}, \bibinfo{person}{Zhiyuan Sun},
  \bibinfo{person}{Xingdi Yuan}, \bibinfo{person}{Marc-Alexandre C{\^o}t{\'e}},
  {and} \bibinfo{person}{Bang Liu}.} \bibinfo{year}{2024}\natexlab{}.
\newblock \showarticletitle{{OPE}x: A Component-Wise Analysis of {LLM}-Centric
  Agents in Embodied Instruction Following}. In
  \bibinfo{booktitle}{\emph{Proceedings of the 62nd Annual Meeting of the
  Association for Computational Linguistics (Volume 1: Long Papers)}},
  \bibfield{editor}{\bibinfo{person}{Lun-Wei Ku}, \bibinfo{person}{Andre
  Martins}, {and} \bibinfo{person}{Vivek Srikumar}} (Eds.).
  \bibinfo{publisher}{Association for Computational Linguistics},
  \bibinfo{address}{Bangkok, Thailand}, \bibinfo{pages}{622--636}.
\newblock
\urldef\tempurl%
\url{https://doi.org/10.18653/v1/2024.acl-long.37}
\showDOI{\tempurl}


\bibitem[\protect\citeauthoryear{Shimojima}{Shimojima}{2015}]%
        {Shimojima2015-SHISPO-2}
\bibfield{author}{\bibinfo{person}{Atsushi Shimojima}.}
  \bibinfo{year}{2015}\natexlab{}.
\newblock \bibinfo{booktitle}{\emph{Semantic Properties of Diagrams and Their
  Cognitive Potentials}}.
\newblock \bibinfo{publisher}{CSLI Publications}, \bibinfo{address}{Stanford,
  California}.
\newblock


\bibitem[\protect\citeauthoryear{S{\o}gaard}{S{\o}gaard}{2023}]%
        {sogaard2023grounding}
\bibfield{author}{\bibinfo{person}{Anders S{\o}gaard}.}
  \bibinfo{year}{2023}\natexlab{}.
\newblock \showarticletitle{Grounding the vector space of an octopus: Word
  meaning from raw text}.
\newblock \bibinfo{journal}{\emph{Minds and Machines}} \bibinfo{volume}{33},
  \bibinfo{number}{1} (\bibinfo{year}{2023}), \bibinfo{pages}{33--54}.
\newblock


\bibitem[\protect\citeauthoryear{Suchan, Bhatt, and Jhavar}{Suchan
  et~al\mbox{.}}{2015}]%
        {DBLP:journals/corr/SuchanBJ15}
\bibfield{author}{\bibinfo{person}{Jakob Suchan}, \bibinfo{person}{Mehul
  Bhatt}, {and} \bibinfo{person}{Harshita Jhavar}.}
  \bibinfo{year}{2015}\natexlab{}.
\newblock \showarticletitle{Talking about the Moving Image: {A} Declarative
  Model for Image Schema Based Embodied Perception Grounding and Language
  Generation}.
\newblock \bibinfo{journal}{\emph{CoRR}}  \bibinfo{volume}{abs/1508.03276}
  (\bibinfo{year}{2015}).
\newblock
\showeprint[arXiv]{1508.03276}
\urldef\tempurl%
\url{http://arxiv.org/abs/1508.03276}
\showURL{%
\tempurl}


\bibitem[\protect\citeauthoryear{Tamari, Shani, Hope, Petruck, Abend, and
  Shahaf}{Tamari et~al\mbox{.}}{2020}]%
        {tamari-etal-2020-language}
\bibfield{author}{\bibinfo{person}{Ronen Tamari}, \bibinfo{person}{Chen Shani},
  \bibinfo{person}{Tom Hope}, \bibinfo{person}{Miriam R~L Petruck},
  \bibinfo{person}{Omri Abend}, {and} \bibinfo{person}{Dafna Shahaf}.}
  \bibinfo{year}{2020}\natexlab{}.
\newblock \showarticletitle{{L}anguage (Re)modelling: {T}owards Embodied
  Language Understanding}. In \bibinfo{booktitle}{\emph{Proceedings of the 58th
  Annual Meeting of the Association for Computational Linguistics}},
  \bibfield{editor}{\bibinfo{person}{Dan Jurafsky}, \bibinfo{person}{Joyce
  Chai}, \bibinfo{person}{Natalie Schluter}, {and} \bibinfo{person}{Joel
  Tetreault}} (Eds.). \bibinfo{publisher}{Association for Computational
  Linguistics}, \bibinfo{address}{Online}, \bibinfo{pages}{6268--6281}.
\newblock
\urldef\tempurl%
\url{https://doi.org/10.18653/v1/2020.acl-main.559}
\showDOI{\tempurl}


\bibitem[\protect\citeauthoryear{Wachowiak and Gromann}{Wachowiak and
  Gromann}{2022}]%
        {wachowiak-gromann-2022-systematic}
\bibfield{author}{\bibinfo{person}{Lennart Wachowiak} {and}
  \bibinfo{person}{Dagmar Gromann}.} \bibinfo{year}{2022}\natexlab{}.
\newblock \showarticletitle{Systematic Analysis of Image Schemas in Natural
  Language through Explainable Multilingual Neural Language Processing}. In
  \bibinfo{booktitle}{\emph{Proceedings of the 29th International Conference on
  Computational Linguistics}}, \bibfield{editor}{\bibinfo{person}{Nicoletta
  Calzolari}, \bibinfo{person}{Chu-Ren Huang}, \bibinfo{person}{Hansaem Kim},
  \bibinfo{person}{James Pustejovsky}, \bibinfo{person}{Leo Wanner},
  \bibinfo{person}{Key-Sun Choi}, \bibinfo{person}{Pum-Mo Ryu},
  \bibinfo{person}{Hsin-Hsi Chen}, \bibinfo{person}{Lucia Donatelli},
  \bibinfo{person}{Heng Ji}, \bibinfo{person}{Sadao Kurohashi},
  \bibinfo{person}{Patrizia Paggio}, \bibinfo{person}{Nianwen Xue},
  \bibinfo{person}{Seokhwan Kim}, \bibinfo{person}{Younggyun Hahm},
  \bibinfo{person}{Zhong He}, \bibinfo{person}{Tony~Kyungil Lee},
  \bibinfo{person}{Enrico Santus}, \bibinfo{person}{Francis Bond}, {and}
  \bibinfo{person}{Seung-Hoon Na}} (Eds.). \bibinfo{publisher}{International
  Committee on Computational Linguistics}, \bibinfo{address}{Gyeongju, Republic
  of Korea}, \bibinfo{pages}{5571--5581}.
\newblock
\urldef\tempurl%
\url{https://aclanthology.org/2022.coling-1.493/}
\showURL{%
\tempurl}


\bibitem[\protect\citeauthoryear{Wałęga, Schultz, and Bhatt}{Wałęga
  et~al\mbox{.}}{2017}]%
        {walega_schultz_bhatt_2017}
\bibfield{author}{\bibinfo{person}{Przemysław~Andrzej Wałęga},
  \bibinfo{person}{Carl Schultz}, {and} \bibinfo{person}{Mehul Bhatt}.}
  \bibinfo{year}{2017}\natexlab{}.
\newblock \showarticletitle{Non-monotonic spatial reasoning with answer set
  programming modulo theories}.
\newblock \bibinfo{journal}{\emph{Theory and Practice of Logic Programming}}
  \bibinfo{volume}{17}, \bibinfo{number}{2} (\bibinfo{year}{2017}),
  \bibinfo{pages}{205–225}.
\newblock
\urldef\tempurl%
\url{https://doi.org/10.1017/S1471068416000193}
\showDOI{\tempurl}


\bibitem[\protect\citeauthoryear{Wicke and Wachowiak}{Wicke and
  Wachowiak}{2024}]%
        {wicke-wachowiak-2024-exploring}
\bibfield{author}{\bibinfo{person}{Philipp Wicke} {and}
  \bibinfo{person}{Lennart Wachowiak}.} \bibinfo{year}{2024}\natexlab{}.
\newblock \showarticletitle{Exploring Spatial Schema Intuitions in Large
  Language and Vision Models}. In \bibinfo{booktitle}{\emph{Findings of the
  Association for Computational Linguistics: ACL 2024}},
  \bibfield{editor}{\bibinfo{person}{Lun-Wei Ku}, \bibinfo{person}{Andre
  Martins}, {and} \bibinfo{person}{Vivek Srikumar}} (Eds.).
  \bibinfo{publisher}{Association for Computational Linguistics},
  \bibinfo{address}{Bangkok, Thailand}, \bibinfo{pages}{6102--6117}.
\newblock
\urldef\tempurl%
\url{https://doi.org/10.18653/v1/2024.findings-acl.365}
\showDOI{\tempurl}


\bibitem[\protect\citeauthoryear{Yang, Xiong, Payani, Shareghi, and Fekri}{Yang
  et~al\mbox{.}}{2024}]%
        {yang-etal-2024-harnessing}
\bibfield{author}{\bibinfo{person}{Yuan Yang}, \bibinfo{person}{Siheng Xiong},
  \bibinfo{person}{Ali Payani}, \bibinfo{person}{Ehsan Shareghi}, {and}
  \bibinfo{person}{Faramarz Fekri}.} \bibinfo{year}{2024}\natexlab{}.
\newblock \showarticletitle{Harnessing the Power of Large Language Models for
  Natural Language to First-Order Logic Translation}. In
  \bibinfo{booktitle}{\emph{Proceedings of the 62nd Annual Meeting of the
  Association for Computational Linguistics (Volume 1: Long Papers)}},
  \bibfield{editor}{\bibinfo{person}{Lun-Wei Ku}, \bibinfo{person}{Andre
  Martins}, {and} \bibinfo{person}{Vivek Srikumar}} (Eds.).
  \bibinfo{publisher}{Association for Computational Linguistics},
  \bibinfo{address}{Bangkok, Thailand}, \bibinfo{pages}{6942--6959}.
\newblock
\urldef\tempurl%
\url{https://doi.org/10.18653/v1/2024.acl-long.375}
\showDOI{\tempurl}


\end{thebibliography}


\end{document}